\documentclass[runningheads]{llncs}
\usepackage{graphicx}

\usepackage{tikz}
\usepackage{comment}
\usepackage{amsmath,amssymb} 
\usepackage{color}
\usepackage{orcidlink}

\usepackage{pifont}
\usepackage{lipsum}
\usepackage{dsfont}
\usepackage{array}
\usepackage{xr}
\usepackage[nice]{nicefrac}
\usepackage{float}
\usepackage{makecell}
\usepackage{graphicx}
\usepackage{xspace}

\newcommand{\normres}[2]{\ensuremath{#1} \footnotesize{(\ensuremath{#2})}}
\newcommand{\boldres}[2]{\ensuremath{\boldsymbol{#1}} \footnotesize{(\ensuremath{\boldsymbol{#2}})}}

\newcommand{\methodname}{Transfer without Forgetting\xspace}
\newcommand{\methnam}{TwF\xspace}
\newcommand{\dpp}{DER\texttt{++}\xspace}

\newcommand{\loss}[1]{\mathcal{L}_{\operatorname{#1}}}

\newcommand{\rebres}[1]{\small{#1}}

\newcommand{\cmark}{\ding{51}}%
\newcommand{\xmark}{\ding{55}}%

\usepackage[accsupp]{axessibility}  

\begin{document}
\pagestyle{headings}
\mainmatter
\def\ECCVSubNumber{7950}  

\title{Transfer without Forgetting}

\titlerunning{Transfer without Forgetting}
%
\author{Matteo Boschini\inst{1}\orcidlink{0000-0002-2809-813X} \and
Lorenzo Bonicelli\inst{1}\orcidlink{0000-0002-9717-5602} \and
Angelo Porrello\inst{1}\orcidlink{0000-0002-9022-8484}  \and \\
Giovanni Bellitto\inst{2}\orcidlink{0000-0002-1333-8348}  \and
Matteo Pennisi\inst{2}\orcidlink{0000-0002-6721-4383}  \and
Simone Palazzo\inst{2}\orcidlink{0000-0002-2441-0982}  \and \\
Concetto Spampinato\inst{2}\orcidlink{0000-0001-6653-2577} \and
Simone Calderara\inst{1}\orcidlink{0000-0001-9056-1538}}
\authorrunning{M. Boschini et al.}
%
\institute{AImageLab, University of Modena and Reggio Emilia, Italy \email{firstname.lastname@unimore.it}\\ \and
PeRCeiVe Lab, University of Catania, Italy \\
\email{firstname.lastname@unict.it}}
\maketitle

\begin{abstract}
This work investigates the entanglement between Continual Learning (CL) and Transfer Learning (TL). In particular, we shed light on the widespread application of network pretraining, highlighting that it is itself subject to catastrophic forgetting. Unfortunately, this issue leads to the under-exploitation of knowledge transfer during later tasks. On this ground, we propose \methodname (\methnam), a hybrid approach building upon a fixed pretrained sibling network, which continuously propagates the knowledge inherent in the source domain through a layer-wise loss term. Our experiments indicate that \methnam steadily outperforms other CL methods across a variety of settings, averaging a 4.81\% gain in Class-Incremental accuracy over a variety of datasets and different buffer sizes. Our code is available at \url{https://github.com/mbosc/twf}.
\end{abstract}

\section{Introduction}
Thanks to the enthusiastic development carried out by the scientific community, there exist myriad widely available deep learning models that can be either readily deployed or easily adapted to perform complex tasks~\cite{he2015delving,silver2016mastering,vinyals2019grandmaster,porrello2019spotting,allegretti2021supporting}. However, the desiderata of practical applications~\cite{shaheen2022continual} often overstep the boundaries of the typical \textit{i.i.d.} paradigm, fostering the study of different learning approaches.

In contrast with the natural tendency of biological intelligence to seamlessly acquire new skills and notions, deep models are prone to an issue known as \textit{catastrophic forgetting}~\cite{mccloskey1989catastrophic}, \textit{i.e.}, they fit the current input data distribution to the detriment of previously acquired knowledge. In light of this limitation, the sub-field of Continual Learning (CL)~\cite{de2019continual,parisi2019continual,van2019three} aspires to train models capable of adaptation and lifelong learning when facing a sequence of changing tasks, either through appositely designed architectures~\cite{rusu2016progressive,schwarz2018progress,mallya2018packnet}, targeted regularization~\cite{li2017learning,kirkpatrick2017overcoming,zenke2017continual} or by storing and replaying previous data points~\cite{rebuffi2017icarl,riemer2018learning,buzzega2020dark,chaudhry2019tiny}.

On a similar note, human intelligence is especially versatile in that it excels in contrasting and incorporating knowledge coming from multiple domains. Instead, the application of deep supervised learning algorithms typically demands large annotated datasets, whose collection has significant costs and may be impractical. To address this issue, Transfer Learning (TL) techniques are typically applied with the purpose of transferring and re-using knowledge across different data domains. In this setting, the simplest technique is to pretrain the model on a huge labeled dataset (\textit{i.e.} the source) and then finetune it on the \textit{target} task~\cite{ren2015faster,he2017mask,devlin2018bert}. Such a simple schema has been recently overcome by more sophisticated domain adaptation algorithms~\cite{chen2020simple,long2017deep,long2018conditional} mainly based on the concept of \textit{feature alignment}: here, the goal is to reduce the shift between the feature distributions of target and source domains. Unfortunately, these approaches often require the availability of the source dataset during training, which clashes with the usual constraints imposed in the CL scenarios.

In this work, we explore the interactions between pretraining and CL and highlight a blind spot of continual learners. Previous work underlined that naive pretraining is beneficial as it leads the learner to reduced forgetting~\cite{mehta2021empirical}. However, we detect that the pretraining task itself is swiftly and catastrophically forgotten as the model veers towards the newly introduced stream of data. This matter is not really detrimental if all target classes are available at once (\textit{i.e.}, joint training): as their exemplars can be accessed simultaneously, the learner can discover a joint feature alignment that works well for all of them while leaving its pretraining initialization. However, if classes are shown in a sequential manner, we argue that transfer mostly concerns the early encountered tasks: as a consequence, pretraining ends up being fully beneficial only for the former classes. For the later ones, since pretraining features are swiftly overwritten, the benefit of pretraining is  instead lowered, thus undermining the advantages of the source knowledge. In support of this argument, this work reports several experimental analyses (Sec.~\ref{sec:pretrandcl}) revealing that state-of-the-art CL methods do not take full advantage of pretraining knowledge.

To account for such a disparity and let all tasks profit equally from pretraining, this work sets up a framework based on Transfer Learning techniques. We show that the Continual Learning setting requires specific and \textit{ad-hoc} strategies to fully exploit the source knowledge without incurring its forgetting. Consequently, we propose an approach termed \textbf{\methodname (\methnam)} that equips the base model with a pretrained and fixed sibling network, which continuously propagates its internal representations to the former network through a per-layer strategy based on \textit{knowledge distillation}~\cite{hinton2015distilling}. We show that our proposal is more effective than alternative approaches (\textit{i.e.}, extending anti-forgetting regularization to the pretraining initialization) and beneficial even if the data used for pretraining is strongly dissimilar w.r.t.\ to the target task.
\section{Related Work}
Continual Learning (CL)~\cite{de2019continual,parisi2019continual} is an increasingly popular field of machine learning that deals with the mitigation of catastrophic forgetting~\cite{mccloskey1989catastrophic}. CL methods are usually grouped as follows, according to the approach they take.

\textit{Regularization-based} methods~\cite{kirkpatrick2017overcoming,lopez2017gradient,chaudhry2019efficient,chaudhry2018riemannian} typically identify subsets of weights that are highly functional for the representations of previous tasks, with the purpose to prevent their drastic modification through apposite optimization constraints. Alternatively, they consolidate the previous knowledge by using past models as soft teachers while learning the current task~\cite{li2017learning}.

\textit{Architectural} approaches dedicate distinct sets of parameters to each task, often resorting to network expansion as new tasks arrive~\cite{rusu2016progressive,mallya2018packnet,serra2018overcoming}. While capable of high performance, they are mostly limited to the Task-IL scenario (described in Sec.~\ref{sec:exp_sett}) as they require task-identifiers at inference time.

\textit{Rehearsal-based} methods employ a fixed-size buffer to store a fraction of the old data. ER~\cite{ratcliff1990connectionist,robins1995catastrophic} interleaves training samples from the current task with previous samples: notably, several works~\cite{farquhar2018towards,buzzega2020dark} point out that such a simple strategy can effectively mitigate forgetting and achieve superior performance. This method has hence inspired several works: DER~\cite{buzzega2020dark} and its extension X-DER~\cite{boschini2022class} also store past model responses and pin them as an additional teaching signal. MER~\cite{riemer2018learning} combines replay and meta-learning~\cite{finn2017model,nichol2018reptile} to maximize transfer from the past while minimizing interference. Other works~\cite{aljundi2019gradient,buzzega2020rethinking} propose different sample-selection strategies to include in the buffer, while GEM~\cite{lopez2017gradient} and its relaxation A-GEM~\cite{chaudhry2019efficient} employ old training data to minimize interference. On a final note, recent works~\cite{boschini2021continual,smith2021memory} exploit the memory buffer to address semi-supervised settings where examples can be either labeled or not.

\noindent \textbf{Transfer Learning (TL)} \cite{pan2009survey} is a machine learning methodology aiming at using the knowledge acquired on a prior task to solve a distinct target task. In its classical formulation~\cite{yosinski2014transferable}, a model is trained on the source dataset and then finetuned on the (possibly much smaller) target dataset to adapt the previously learned features. Alternatively, transfer can be induced via multi-level Knowledge Distillation, guided by meta-learning~\cite{jang2019learning}, attention~\cite{wang2019pay} or higher-level descriptions of the flow of information within the model~\cite{yim2017gift}.
\section{Method}
\subsubsection{Setting.} In CL, a classification model $f_{(\theta, \phi)}$ (composed of a multilayered feature extractor $h_\theta=h_{\theta_l}^{(l)}\circ h_{\theta_{l-1}}^{(l-1)}\circ \dots \circ h_{\theta_1}^{(1)}$ and a classifier $g_\phi$, $f_{(\theta, \phi)} = g_\phi \circ h_\theta$) is trained on a sequence of $N$ tasks $\mathcal{T}_i = \{(x_j^i,y_j^i)\}_{j = 1}^{|\mathcal{T}_i|}$. The objective of $f_{(\theta,\phi)}$ is minimizing the classification error across all seen tasks:
\begin{equation}
\label{eq:cont_obj}
    \operatornamewithlimits{min}_{\theta,\phi} \mathcal{L} =
    \mathop{\mathbb{E}}_{i} \bigg[ \mathop{\mathbb{E}}_{(x,y)\sim \mathcal{T}_i} \Big[ \ell(y, f_{(\theta, \phi)}(x)) \Big] \bigg],
\end{equation}
\noindent where $\ell$ is a suitable loss function. Unfortunately, the problem framed by Eq.~\ref{eq:cont_obj} cannot be directly optimized due to the following key assumptions: \textit{i)} while learning the current task $\mathcal{T}_c$, examples and labels of previous tasks are inaccessible; \textit{ii)} the label space of distinct tasks is disjoint ($y^i_m \neq y^j_n ~ \forall i \neq j$) \textit{i.e.}, classes learned previously cannot recur in later phases. Therefore, Eq.~\ref{eq:cont_obj} can only be approximated, seeking adequate performance on previously seen tasks (\textit{stability}), while remaining flexible enough to adapt to upcoming data (\textit{plasticity}).
\subsection{Pretraining incurs Catastrophic Forgetting}
\label{sec:pretrandcl}
\begin{figure}[t]
    \centering
    \includegraphics[width=\linewidth]{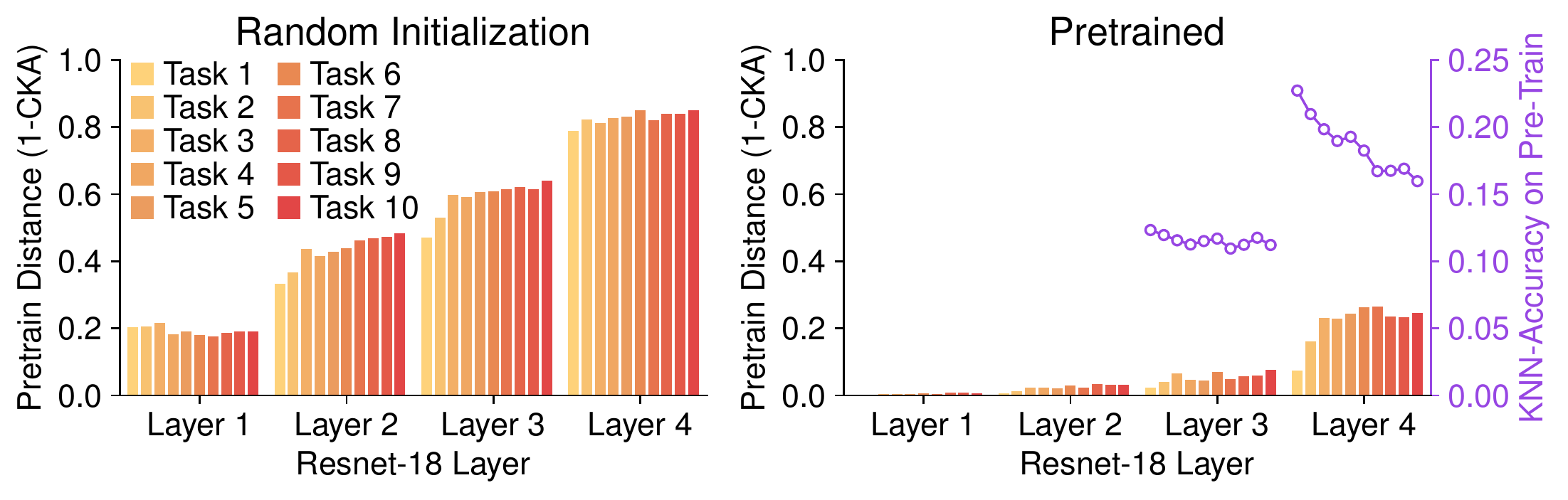}
    \caption{Forgetting of the initialization, measured as the distance from the pretrain ($1-$CKA~\cite{kornblith2019similarity}) (lower is better) and $k$NN accuracy (higher is better). 
    Features extracted by a pretrained model remain closer to the initialization w.r.t.\ a randomly initialized model. Furthermore, the steady decrease in $k$NN accuracy as training progresses reveals that features become less specific for past tasks.}
    \label{fig:forgetting}
\end{figure}
\textit{Mehta et al.}~\cite{mehta2021empirical} have investigated the entanglement between continual learning and pretraining, highlighting that the latter leads the optimization towards wider minima of the loss landscape. As deeply discussed in~\cite{buzzega2020dark,boschini2022class}, such property is strictly linked to a reduced tendency in incurring forgetting.

On this latter point, we therefore provide an alternate experimental proof of the benefits deriving from pretraining initialization. In particular, we focus on ResNet-18 trained with ER~\cite{robins1995catastrophic} on Split CIFAR-100\footnote{This preliminary experiment follows the same setting presented in Sec.~\ref{sec:exp_sett}} and measure how each individual layer differs from its initialization. 
It can be observed that a randomly initialized backbone (Fig.~\ref{fig:forgetting}, \textit{left}) significantly alters its parameters at all layers while tasks progress, resulting in a very low Centered Kernel Alignment~\cite{kornblith2019similarity} similarity score already at the first CL task. On the contrary, a backbone pretrained on Tiny ImageNet (Fig.~\ref{fig:forgetting}, \textit{right}) undergoes limited parameter variations in its layers, with the exception of the last residual layer (although to a lesser extent w.r.t.\ random init.).
This latter finding indicates that its pretraining parametrization requires relevant modifications to fit the current training data. This leads to the \textit{catastrophic forgetting} of the source pretraining task: namely, the latter is swiftly forgotten as the network focuses on the initial CL tasks. This is corroborated by the decreasing accuracy for pretraining data of a $k$NN classifier trained on top of \textit{Layer 3} and \textit{Layer 4} representations in Fig.~\ref{fig:forgetting} (\textit{right}).

To sum up, while pretraining is certainly beneficial, the model drifts away from it one task after the other. Hence, only the first task takes full advantage of it; the optimization of later tasks, instead, starts from an initialization that increasingly differs from the one attained by pretraining. This is detrimental, as classes introduced later might be likewise advantaged by the reuse of different pieces of the initial knowledge.
\subsection{\methodname}
\begin{figure}[t]
    \centering
    \includegraphics[clip, trim=7.04cm 2.09cm 2.44cm 1.80cm, width=\linewidth,keepaspectratio]{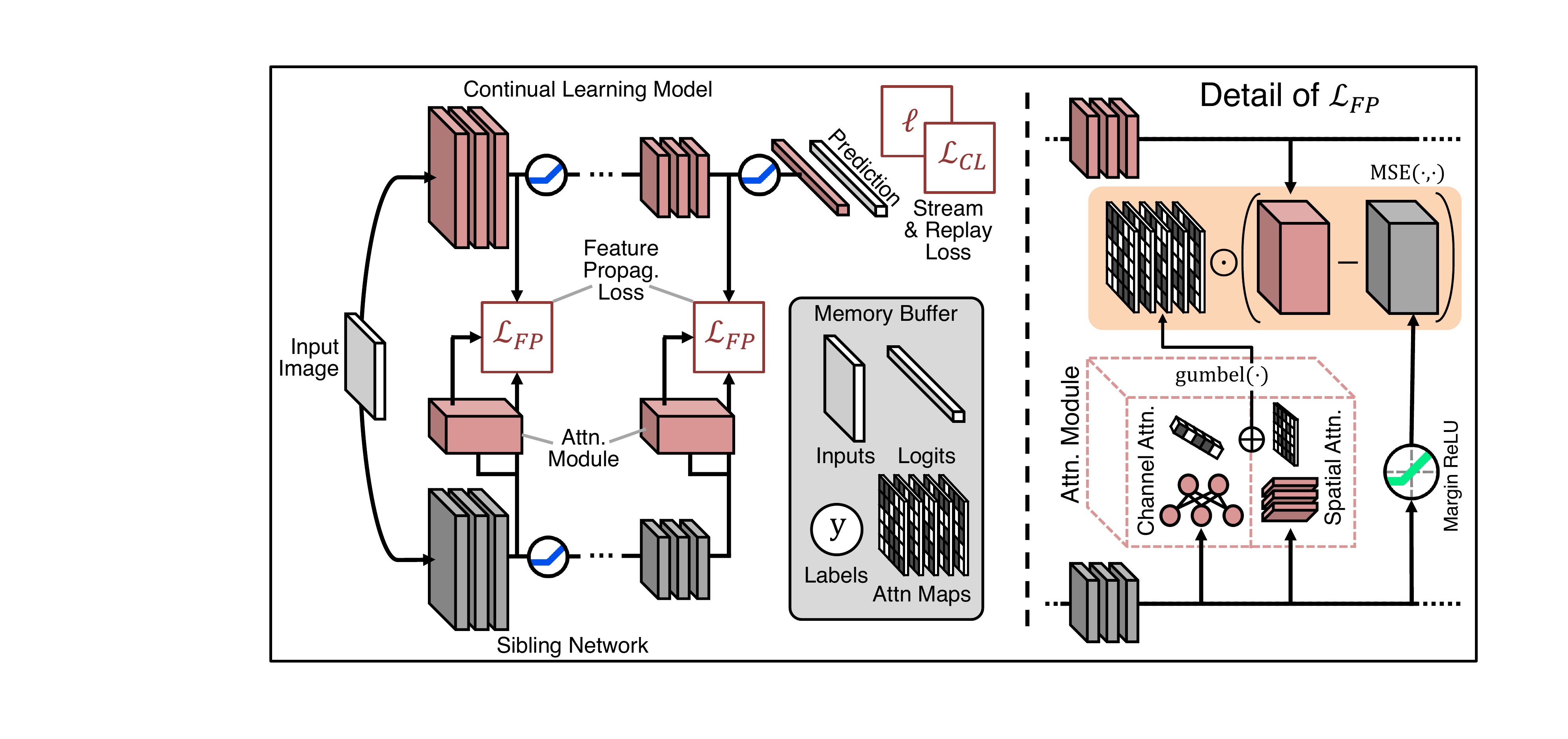}
    \caption{Overview of \methnam and detail of $\mathcal{L}_\text{FP}$: Given a batch of samples from the current task or from $\mathcal{B}$, we \textit{i)} extract intermediate features from both the student and fixed sibling backbones at multiple layers; \textit{ii)} compute the corresponding binarized attention maps $\mathbb{M}(\cdot)$; \textit{iii)} pull the attention-masked representations of the two models closer.}
    \label{fig:model_illustration}
\end{figure}
To mitigate the issue above, we propose a strategy that enables a continuous transfer between the source task and the incrementally learned target problem.
\subsubsection{Feature Propagation.} As the training progresses, the input stream introduces new classes that might benefit from the adaptation of specific features of the pretrained model. To enable feature transfer without incurring pretraining forgetting, we maintain a copy of it (the \textit{sibling} model) and adopt an intermediate feature knowledge distillation~\cite{romero2014fitnets,aguilar2020knowledge,wang2019pay,heo2019comprehensive,monti2022many} objective. Considering a subset of $L$ layers, we seek to minimize the distance between the activations of the base network $h^{(l)}_{\theta} \triangleq h^{(l)}_{\theta}(x)$ and those from its pretrained sibling $\widehat{h}^{(l)} \triangleq h^{(l)}_{\theta^t}(x)$:
\begin{equation}
    \label{eq:obj_1}
    \mathop{\mathbb{E}}_{x\sim \mathcal{T}_c} \bigg[ \sum_{l=1}^L ||{h}^{(l)}_{\theta} - \operatorname{ReLU_\text{m}}(\widehat{h}^{(l)}) ||_2^2 \bigg],
\end{equation}
\noindent where $c$ is the current task and $\operatorname{ReLU_\text{m}}(\cdot)$ indicates the application of a margin ReLU activation~\cite{heo2019comprehensive}. It is noted that the objective outlined by Eq.~\ref{eq:obj_1} leads the CL model to focus on mirroring the internal representations of the pretrained teacher and maximizing transfer. However, focusing on the latter solely can lead to excessive rigidity, thus preventing the model from fitting the data from the current task altogether. On these grounds, we take inspiration from~\cite{wang2019pay} and use a weighted version of Eq.~\ref{eq:obj_1}. In particular, an apposite learnable module computes a gating attention map $\mathbb{M}(\cdot)$ over the feature maps of the sibling, which serves as a binary mask selecting which spatial regions have to be aligned. The resulting objective is consequently updated as follows:
\begin{equation}
    \label{eq:obj_2}
    \mathop{\mathbb{E}}_{x\sim \mathcal{T}_c} \bigg[ \sum_{l=1}^L ||\mathbb{M}(\widehat{h}^{(l)}) \odot \Big( h^{(l)}_{\theta}-{\operatorname{ReLU_\text{m}}}(\widehat{h}^{(l)}) \Big)||_2^2 \bigg],
\end{equation}
\noindent where $\odot$ indicates the Hadamard product between two tensors of the same dimensions. 
The attention maps $\mathbb{M}(\cdot)$ are computed through specific layers, whose architectural design follows the insights provided in~\cite{park2018bam}. Specifically, they forward the input activation maps into two parallel branches, producing respectively a Channel Attention $\mathbb{M}_{\operatorname{Ch}}(\cdot)$ map and a Spatial Attention $\mathbb{M}_{\operatorname{Sp}}(\cdot)$ map. These two intermediate results are summed and then activated through a binary Gumbel-Softmax sampling~\cite{jang2016categorical}, which allows us to model discrete \textit{on-off} decisions regarding which information we want to propagate. In formal terms: 
\begin{equation}
    \label{eq:bam}
    \mathbb{M}(\widehat{h}^{(l)}) \triangleq \operatorname{gumbel}(\mathbb{M}_{\operatorname{Ch}}(\widehat{h}^{(l)}) + \mathbb{M}_{\operatorname{Sp}}(\widehat{h}^{(l)})).
\end{equation}
The Spatial Attention $\mathbb{M}_{\operatorname{Sp}}(\widehat{h}^{(l)})$ regulates the propagation of spatially localized information and is obtained by stacking four convolutional layers~\cite{park2018bam} with different configurations (\textit{i.e.}, the kernel sizes and dilation rates -- please refer to supplementary materials for additional details):
\begin{equation}
    \label{eq:ms}
    \mathbb{M}_{\operatorname{Sp}}(\widehat{h}^{(l)}) \triangleq \operatorname{C}_{1\times1} \circ \operatorname{C}_{3\times3} \circ \operatorname{C}_{3\times3} \circ \operatorname{C}_{1\times1}(\widehat{h}^{(l)}),
\end{equation}
where $\operatorname{C}$ denotes a sequence of convolutional, batch normalization, and ReLU activation layers. On the other hand, the Channel Attention $\mathbb{M}_{\operatorname{Ch}}(\widehat{h}^{(l)})$ estimates the information across the channels of $\widehat{h}^{(l)}$; in its design, we draw inspiration from the formulation proposed in~\cite{ilse2018attention}. Formally, considering the result $\widehat{h}_{\operatorname{GAP}}^{(l)}$ of the Global Average Pooling (GAP) applied on top of $\widehat{h}^{(l)}$, we have:
\begin{equation}
    \label{eq:mc}
    \mathbb{M}_{\operatorname{Ch}}(\widehat{h}^{(l)}) \triangleq \operatorname{tanh}(\operatorname{BN}(W_1^{\mathsf{T}} \widehat{h}_{\operatorname{GAP}}^{(l)})) \cdot \sigma(\operatorname{BN}(W_2^{\mathsf{T}}\widehat{h}_{\operatorname{GAP}}^{(l)})) + W_3^{\mathsf{T}}\widehat{h}_{\operatorname{GAP}}^{(l)},
\end{equation}
\noindent where $W_1$, $W_2$, and $W_3$ are the weights of three fully connected layers organized in parallel and $\operatorname{BN}$ indicates the application of batch normalization.
\subsubsection{Diversity loss.} Without a specific loss term supervising the attention maps, we could incur in useless behaviors, \textit{e.g.}, all binary gates being either on or off, or some channels being always propagated and some others not. While recent works provide a target expected activation ratio~\cite{abati2020conditional,serra2018overcoming} as a countermeasure, we encourage the auxiliary modules to assign different propagation gating masks to different examples. The intuition is that each example has its own preferred subset of channels to be forwarded from the sibling. To do so, we include an additional auxiliary loss term~\cite{muller2020subclass} as follows:
\begin{equation}
\label{eq:diverse}
\begin{split}
    \mathcal{L}_{\operatorname{AUX}} &\triangleq - \lambda \sum_{l=1}^L \mathop{\mathbb{E}}_{x_1, \dots, x_n \sim \mathcal{T}_c} \bigg[ \sum_{j=1}^{n}\log\frac{e^{{g}_{ij}^{\text%
{T}}{g}_{ij}/T}}{\frac{1}{n}\sum_{k=1}^{n}e^{{g}_{ij}^{%
\text{T}}{g}_{ik}/T}} \bigg],\\
g_{ij} &\triangleq \operatorname{NORM}(\operatorname{GAP}(\mathbb{M}(\widehat{h}^{(l)}(x_j)))),
\end{split}
\end{equation}
\noindent where $n$ indicates the batch size, $\operatorname{NORM}$ a normalization layer, $T$ a temperature and finally $\lambda$ is a scalar weighting the contribution of this loss term to the overall objective. In practice, we ask each vector containing channel-wise average activity to have a low dot product with vectors of other examples. 
\subsection{Knowledge Replay}
The training objective of Eq.~\ref{eq:obj_2} is devised to facilitate selective feature transfer between the in-training model and the immutable sibling. However, to prevent forgetting tied to previous CL tasks to the greatest extent, the model should also be provided with a targeted strategy. We thus equip the continual learner with a small memory buffer $\mathcal{B}$ (populated with examples from the input stream via \textit{reservoir sampling}~\cite{vitter1985random}) and adopt the simple labels and logits replay strategy proposed in~\cite{buzzega2020dark}:
\begin{equation}
    \mathcal{L}_{\text{CL}} \triangleq \mathop{\mathbb{E}}_{(x,y,l)\sim \mathcal{B}} \bigg[ \alpha \cdot ||f_{(\theta, \phi)}(x) - l ||_2^2 + \beta \cdot \ell(y, f_{(\theta, \phi)}(x)) \bigg],
\end{equation}
\noindent where $(x,y,l)$ is a triplet of example, label and original network responses $l = f(x)$ recorded at the time of sampling and $\alpha$, $\beta$ are scalar hyperparameters. 
Although extremely beneficial, we remark that the model need not optimize $\mathcal{L}_{\text{CL}}$ to achieve basic robustness against catastrophic forgetting (as shown in Sec.~\ref{sec:ablas}): preserving pretraining features already serves this purpose.

\subsubsection{Replaying past propagation masks.} With the purpose of protecting the feature propagation formulated in Eq.~\ref{eq:obj_2} from forgetting, we also extend it to replay examples stored in memory. It must be noted that doing so requires taking additional steps to prevent cross-task interference; indeed, simply applying Eq.~\ref{eq:obj_2} to replay items would apply the feature propagation procedure unchanged to all tasks, regardless of the classes thereby included. For this reason, we take an extra step and make all batch normalization and fully connected layers in Eq.~\ref{eq:bam}, \ref{eq:ms} and \ref{eq:mc} conditioned~\cite{de2017modulating} w.r.t.\ the CL task.
Consequently, we add to $\mathcal{B}$ for each example $x$ both its task label $t$ and its corresponding set of binary attention maps $m = (m^1, ..., m^l)$ generated at the time of sampling. Eq.~\ref{eq:obj_2} is finally updated as:
\begin{equation}
    \label{eq:obj_2b}
    \begin{split}
    \mathcal{L}_\text{FP} \triangleq& \mathop{\mathbb{E}}_{\substack{(x, t=c) \sim \mathcal{T}_c\\(x;t) \sim \mathcal{B}}} \bigg[ \sum_{l=1}^L ||\mathbb{M}(\widehat{h}^{(l)}; t) \odot \Big(h^{(l)} - \operatorname{ReLU_\text{m}}(\widehat{h}^{(l)}) \Big)||_2^2 \bigg]\\
    +& \mathop{\mathbb{E}}_{\substack{(x,t,m) \sim \mathcal{B}\\l=1,\dots,L}} \bigg[ \operatorname{BCE}\Big(\mathbb{M}(\widehat{h}^{(l)}; t), m^{(l)} \Big) \bigg],
    \end{split}
\end{equation}
where the second term is an additional replay contribution distilling past attention maps, with $\operatorname{BCE}$ indicating the binary cross entropy criterion.

\subsubsection{Overall objective.} Our proposal -- dubbed \textbf{\methodname (\methnam)} -- optimizes the following training objective, also summarized in Fig.~\ref{fig:model_illustration}:
\begin{equation}
    \label{eq:overall}
    \operatornamewithlimits{min}_{\theta,\phi} \mathop{\mathbb{E}}_{(x,y)\sim \mathcal{T}_c} \big[ \ell(y_j^i, f_{(\theta, \phi)}(x_j^i)) \big] + \mathcal{L}_\text{CL} + \mathcal{L}_\text{FP} + \mathcal{L}_{\operatorname{AUX}}.
\end{equation}
We remark that: \textit{i)} while \methnam requires keeping a copy of the pretrained model during training, this does not hold at inference time; \textit{ii)} similarly, task labels $t$ are not needed during inference but only while training, which makes \methnam capable of operating under both the Task-IL and Class-IL CL settings~\cite{van2019three};
\textit{iii)} the addition of $t$ and $m$ in $\mathcal{B}$ induces a limited memory overhead: $t$ can be obtained from the stored labels $y$ for typical classification tasks with a fixed number of classes per task, while $m$ is a set of Boolean maps that is robust to moderate re-scaling (as we demonstrate by storing $m$ at half resolution for our experiments in Sec.~\ref{sec:exp}). We finally point out that, as maps $m$ take discrete binary values, one could profit from lossless compression algorithms (such as Run-Length Encoding~\cite{robinson1967results} or LZ77~\cite{ziv1977universal}) and thus store a compressed representation into the memory buffer. We leave the comprehensive investigation of this application to future works.

\section{Experiments}
\label{sec:exp}
\subsection{Experimental Setting}
\label{sec:exp_sett}
\subsubsection{Metrics.} We assess the overall performance of the models in terms of \textit{Final Average Accuracy} (FAA), defined as the average accuracy on all seen classes after learning the last task, and \textit{Final Forgetting}~\cite{chaudhry2018riemannian} (FF), defined as:
\begin{equation}
    \text{FF} \triangleq \frac{1}{T-1} \sum_{i=0}^{T-2}{\max_{t\in\{0,\ldots,T-2\}}\{a_i^t - a_i^{T-1}\}},
\end{equation}
where $a_i^t$ denotes the accuracy on task $\tau_i$ after training on the $t^{\text{th}}$ task.
\subsubsection{Settings.} We report results on two common protocols~\cite{van2019three}: \textit{Task-Incremental Learning} (Task-IL), where the model must learn to classify samples only from within each task, and \textit{Class-Incremental Learning} (Class-IL), where the model must gradually learn the overall classification problem. 
The former scenario is a relaxation of the latter, as it provides the model with the task identifier of each sample at test time; for this reason, we focus our evaluation mainly on the Class-IL protocol, highlighted as a more realistic and challenging benchmark~\cite{farquhar2018towards,aljundi2019gradient}.
\subsubsection{Datasets.} We initially describe a scenario where the transfer of knowledge from the pretrain is facilitated by the similarity between the two distributions. Precisely, we use \textbf{CIFAR-100}~\cite{krizhevsky2009learning} as the pretrain dataset and then evaluate the models on \textbf{Split CIFAR-10}~\cite{zenke2017continual} ($5$ binary tasks) (see Tab.~\ref{table:c10}). In Tab.~\ref{table:c100} we envision a second and more challenging benchmark, which relies on \textbf{Split CIFAR-100}~\cite{zenke2017continual} with the opportunity to benefit from the knowledge previously learned on \textbf{Tiny ImageNet}~\cite{Le2015TinyIV}. Due to the size mismatch between CIFAR-100 and the samples from Tiny ImageNet, we resize the latter to $32\times 32$ during pretraining.
The last scenario (Tab.~\ref{table:cub200}) involves pretraining on ImageNet~\cite{deng2009imagenet} and learning incrementally \textbf{Split CUB-200}~\cite{chaudhry2019efficient,yu2020semantic}, split into $10$ tasks of $20$ classes each. With an average of only $29.97$ images per class and the use of higher-resolution input samples (resized to $224\times 224$), this benchmark is the most challenging. We use ResNet18~\cite{he2016deep} for all experiments involving Split CIFAR-10 and Split CIFAR-100, as in~\cite{rebuffi2017icarl,buzzega2020dark}, while opting for ResNet50 on Split CUB-200. The supplementary materials report other details on the experimental protocols. 
\subsubsection{Competitors.} We focus our comparison on state-of-the-art rehearsal algorithms, as they prevail on most benchmarks in literature~\cite{chaudhry2019tiny,buzzega2020dark,van2019three}.
\begin{itemize}
    \item \textbf{Experience Replay (ER)}~\cite{ratcliff1990connectionist,robins1995catastrophic} is the first embodiment of a rehearsal strategy that features a small memory buffer containing an \textit{i.i.d.}\ view of all the tasks seen so far. During training, data from the stream is complemented with data sampled from the buffer. While this represents the most straightforward use of a memory in a CL scenario, ER remains a strong baseline, albeit with a non-negligible memory footprint.
    \item \textbf{Dark Experience Replay (DER)}~\cite{buzzega2020dark} envisions a self-distillation~\cite{furlanello2018born} constraint on data stored in the memory buffer and represents a simple extension to the basic rehearsal strategy of ER. In this work, we compare against \dpp, which includes both ER and DER objectives.
    \item \textbf{Incremental Classifier and Representation Learning (iCaRL)}~\cite{rebuffi2017icarl} tackle catastrophic forgetting by distilling the responses of the model at the previous task boundary and storing samples that better represent the current task. In addition to simple replay, those \textit{exemplars} are used to compute class-mean prototypes for nearest-neighbor classification.
    \item \textbf{ER with Asymmetric Cross-Entropy (ER-ACE)}~\cite{caccia2022new} recently introduced a method to alleviate class imbalances to ER. The authors obtain a major gain in accuracy by simply separating the cross-entropy contribution of the classes in the current batch and that of the ones in the memory buffer.
    \item \textbf{Contrastive Continual Learning (CO$^2$L)}~\cite{cha2021co2l} proposes to facilitate knowledge transfer from samples stored in the buffer by optimizing a contrastive learning objective, avoiding any potential bias introduced by a cross-entropy objective. To perform classification, a linear classifier needs to be first trained on the exemplars stored in the buffer.
\end{itemize}
In addition, we also include results from two popular regularization methods. \textbf{Online Elastic Weight Consolidation (oEWC)}~\cite{kirkpatrick2017overcoming} penalizes changes on the most important parameters by means of an online estimate of the Fisher Information Matrix evaluated at task boundaries. \textbf{Learning without Forgetting (LwF)}~\cite{li2017learning} includes a distillation target similar to iCaRL but does not store any exemplars. We remark that \textbf{all competitors undergo an initial pretraining phase} prior to CL, thus ensuring a fair comparison.

To gain a clearer understanding of the results, all the experiments include the performance of the upper bound (\textbf{Joint}), obtained by jointly training on all classes in a non-continua fashion. We also report the results of the model obtained by training sequentially on each task (\textbf{Finetune}), \textit{i.e.}, without any countermeasure to forgetting.
\setlength{\tabcolsep}{4pt}
\begin{table}[t]
\begin{center}
\caption{Final Average Accuracy (FAA) [$\uparrow$] and Final Forgetting (FF) [$\downarrow$] on Split CIFAR-10 w.\ pretrain on CIFAR-100.}
\label{table:c10}
\begin{tabular}{l@{\hskip 0.5cm}cc@{\hskip 0.5cm}cc}
\hline\noalign{\smallskip}
\textbf{FAA (FF)} & \multicolumn{4}{c}{\textbf{Split CIFAR-10}~~(\textit{pretr.\ CIFAR-100})}\\
\noalign{\smallskip}
\hline
\noalign{\smallskip}

\textbf{Method} & \multicolumn{2}{c@{\hskip 0.5cm}}{\textbf{Class-IL}} & \multicolumn{2}{c@{\hskip 0.5cm}}{\textbf{Task-IL}}\\
\noalign{\smallskip}
\hline
\noalign{\smallskip}
Joint    (UB)                     & \multicolumn{2}{c}{\normres{92.89}{-}}          & \multicolumn{2}{c}{\normres{98.38}{-}} \\
Finetune                          & \multicolumn{2}{c}{\normres{19.76}{98.11}}      & \multicolumn{2}{c}{\normres{84.05}{17.75}} \\
oEwC~\cite{schwarz2018progress}   & \multicolumn{2}{c}{\normres{26.10}{88.85}}      & \multicolumn{2}{c}{\normres{81.84}{19.50}} \\
LwF~\cite{li2017learning}         & \multicolumn{2}{c}{\normres{19.80}{97.96}}      & \multicolumn{2}{c}{\normres{86.41}{14.35}} \\
\noalign{\smallskip}
\hline
\noalign{\smallskip}
\textbf{Buffer Size} & $500$ & $5120$ & $500$ & $5120$ \\
\noalign{\smallskip}
\hline
\noalign{\smallskip}
ER~\cite{robins1995catastrophic}  & \normres{67.24}{38.24} & \normres{86.27}{13.68} & \normres{96.27}{2.23}  & \normres{97.89}{0.55}\\
CO$^2$L~\cite{cha2021co2l}           & \normres{75.47}{21.80}  & \normres{87.59}{9.61} & \normres{96.77}{1.23} & \normres{97.82}{0.53} \\
iCaRL~\cite{rebuffi2017icarl}     & \normres{76.73}{14.70} & \normres{77.95}{12.90} & \normres{97.25}{0.74}  & \normres{97.52}{0.15}\\
\dpp~\cite{buzzega2020dark}      & \normres{78.42}{20.18} & \normres{87.88}{8.02}  & \normres{94.25}{4.46}  & \normres{96.42}{1.99}\\
ER-ACE~\cite{caccia2022new}       & \normres{77.83}{10.63} & \normres{86.20}{5.58} & \normres{96.41}{2.11} & \normres{97.60}{0.66}\\
\textbf{\methnam (ours)}               & \boldres{83.65}{11.59} & \boldres{89.55}{6.85}  & \boldres{97.49}{0.86}  & \boldres{98.35}{0.17}\\

\hline
\end{tabular}
\end{center}
\end{table}
\setlength{\tabcolsep}{1.4pt}

\subsection{Comparison with State-Of-The-Art}
\setlength{\tabcolsep}{4pt}
\begin{table}[t]
\begin{center}
\caption{Accuracy (forgetting) on Split CIFAR-100 w.\ pretrain on Tiny ImageNet.}
\label{table:c100}
\begin{tabular}{l@{\hskip 0.5cm}cc@{\hskip 0.5cm}cc}
\hline\noalign{\smallskip}
\textbf{FAA (FF)} & \multicolumn{4}{c}{\textbf{Split CIFAR-100}~~(\textit{pretr.\ Tiny ImageNet})}\\
\noalign{\smallskip}
\hline
\noalign{\smallskip}
\textbf{Method} & \multicolumn{2}{c@{\hskip 0.5cm}}{\textbf{Class-IL}} & \multicolumn{2}{c@{\hskip 0.5cm}}{\textbf{Task-IL}}\\
\noalign{\smallskip}
\hline
\noalign{\smallskip}
Joint    (UB)                     & \multicolumn{2}{c}{\normres{75.20}{-}}          & \multicolumn{2}{c}{\normres{93.40}{-}} \\
Finetune                          & \multicolumn{2}{c}{\normres{09.52}{92.31}}      & \multicolumn{2}{c}{\normres{73.50}{20.53}} \\
oEwC~\cite{schwarz2018progress}   & \multicolumn{2}{c}{\normres{10.95}{81.71}}      & \multicolumn{2}{c}{\normres{65.56}{21.33}} \\
LwF~\cite{li2017learning}         & \multicolumn{2}{c}{\normres{10.83}{90.87}}      & \multicolumn{2}{c}{\normres{86.19}{4.77}} \\
\noalign{\smallskip}
\hline
\noalign{\smallskip}
\textbf{Buffer Size} & $500$ & $2000$ & $500$ & $2000$ \\
\noalign{\smallskip}
\hline
\noalign{\smallskip}
ER~\cite{robins1995catastrophic}  & \normres{31.30}{65.40} & \normres{46.80}{46.95} & \normres{85.98}{6.14} & \normres{87.59}{4.85}\\
CO$^2$L~\cite{cha2021co2l}           & \normres{33.40}{45.21} & \normres{50.95}{31.20} & \normres{68.51}{21.51} & \normres{82.96}{8.53} \\
iCaRL~\cite{rebuffi2017icarl}     & \normres{56.00}{19.27} & \normres{58.10}{16.89} & \boldres{89.99}{2.32} & \normres{90.75}{1.68}\\
\dpp~\cite{buzzega2020dark}      & \normres{43.65}{48.72} & \normres{58.05}{29.65} & \normres{73.86}{20.08} & \normres{86.63}{6.86}\\
ER-ACE~\cite{caccia2022new}       & \normres{53.38}{21.63} & \normres{57.73}{17.12} & \normres{87.21}{3.33} & \normres{88.46}{2.46}\\
\textbf{\methnam (ours)}               & \boldres{56.83}{23.89} & \boldres{64.46}{15.23} & \normres{89.82}{3.06} & \boldres{91.11}{2.24}\\

\hline
\end{tabular}
\end{center}
\end{table}
\setlength{\tabcolsep}{1.4pt}

\setlength{\tabcolsep}{4pt}
\begin{table}[t]
\begin{center}
\caption{Accuracy (forgetting) on Split CUB-200 w.\ pretrain on ImageNet.}
\label{table:cub200}
\begin{tabular}{l@{\hskip 0.5cm}cc@{\hskip 0.5cm}cc}
\hline\noalign{\smallskip}
\textbf{FAA (FF)} & \multicolumn{4}{c}{\textbf{Split CUB-200}~~(\textit{pretr.\ ImageNet})}\\
\noalign{\smallskip}
\hline
\noalign{\smallskip}
\textbf{Method} & \multicolumn{2}{c@{\hskip 0.5cm}}{\textbf{Class-IL}} & \multicolumn{2}{c@{\hskip 0.5cm}}{\textbf{Task-IL}}\\
\noalign{\smallskip}
\hline
\noalign{\smallskip}
Joint    (UB)                     & \multicolumn{2}{c}{\normres{78.54}{-}}          & \multicolumn{2}{c}{\normres{86.48}{-}} \\
Finetune                          & \multicolumn{2}{c}{\normres{8.56}{82.38}}      & \multicolumn{2}{c}{\normres{36.84}{50.95}} \\
oEwC~\cite{schwarz2018progress}   & \multicolumn{2}{c}{\normres{8.20}{71.46}}      & \multicolumn{2}{c}{\normres{33.94}{40.36}} \\
LwF~\cite{li2017learning}         & \multicolumn{2}{c}{\normres{8.59}{82.14}}      & \multicolumn{2}{c}{\normres{22.17}{67.08}} \\
\noalign{\smallskip}
\hline
\noalign{\smallskip}
\textbf{Buffer Size} & $400$ & $1000$ & $400$ & $1000$ \\
\noalign{\smallskip}
\hline
\noalign{\smallskip}
ER~\cite{robins1995catastrophic}  & \normres{45.82}{40.76} & \normres{59.88}{25.65} & \normres{75.26}{9.82} & \normres{80.19}{4.52}\\
CO$^2$L~\cite{cha2021co2l}           & \normres{8.96}{32.04} & \normres{16.53}{20.99} & \normres{22.91}{26.42} & \normres{35.79}{16.61} \\
iCaRL~\cite{rebuffi2017icarl}     & \normres{46.55}{12.48} & \normres{49.07}{11.24} & \normres{68.90}{3.14} & \normres{70.57}{3.03}\\
\dpp~\cite{buzzega2020dark}      & \normres{56.38}{26.59} & \normres{67.35}{13.47} & \normres{77.16}{7.74} & \normres{82.00}{3.25}\\
ER-ACE~\cite{caccia2022new}       & \normres{48.18}{25.79} & \normres{58.19}{16.56} & \normres{74.34}{9.78} & \normres{78.27}{6.09}\\
\textbf{\methnam (ours)}               & \boldres{57.78}{18.32} & \boldres{68.32}{6.74} & \boldres{79.35}{5.77} & \boldres{82.81}{2.14}\\

\hline
\end{tabular}
\end{center}
\end{table}
\setlength{\tabcolsep}{1.4pt}

\subsubsection{Regularization methods.} Across the board, non-rehearsal methods (oEWC and LwF) manifest a profound inability to effectively use the features learned during the pretrain. As those methods are not designed to extract and reuse any useful features from the initialization, the latter is rapidly forgotten, thus negating any knowledge transfer in later tasks. This is particularly true for oEWC, whose objective proves to be both too strict to effectively learn the current task and insufficient to retain the initialization. Most notably, on Split CUB-200 oEWC shows performance lower than Finetune on both Task- and Class-IL.

\subsubsection{Rehearsal methods.} In contrast, rehearsal models that feature some form of distillation (\dpp and iCaRL) manage to be competitive on all benchmarks. In particular, iCaRL proves especially effective on Split CIFAR-100, where it reaches the second highest FAA even when equipped with a small memory thanks to its \textit{herding} buffer construction strategy. However, this effect is less pronounced on Split CIFAR-10 and Split CUB-200, where the role of pretraining is far more essential due to the similarity of the two distributions for the former and the higher difficulty of the latter. In these settings, we see iCaRL fall short of \dpp, which better manages to maintain and reuse the features available from its initialization. Moreover, we remark that iCaRL and \dpp show ranging Class-IL performance in different tasks, whereas our method is much less sensitive to the specific task at hand. 

While it proves effective on the easier Split CIFAR-10 benchmark, CO$^2$L does not reach satisfactory results on either Split CIFAR-100 or Split CUB-200. We ascribe this result to the high sensitivity of this model to the specifics of its training process (\textit{e.g.}, to the applied transforms and the number of epochs required to effectively train the feature extractor with a contrastive loss). Remarkably, while we extended the size of the batch in all experiments with CO$^2$L to $256$ to provide a large enough pool of negative samples, it still shows off only a minor improvement on non-rehearsal methods for Split CUB-200.

Interestingly, while both ER and ER-ACE do not feature distillation, we find their performance to be competitive for large enough buffers. In particular, the asymmetric objective of ER-ACE appears less sensitive to a small memory buffer but always falls short of \dpp when this constraint is less severe.

\subsubsection{\methodname.} Finally, results across all proposed benchmarks depict our method (\methnam) as consistently outperforming all the competitors, with an average gain of $4.81\%$ for the Class-IL setting and $2.77\%$ for the Task-IL setting, w.r.t.\ the second-best performer across all datasets (\dpp and ER-ACE, respectively). This effect is especially pronounced for smaller buffers on Split CIFAR-10 and Split CUB-200, for which the pretrain provides a valuable source of knowledge to be transferred. We argue that this proves the efficacy of our proposal to retain and adapt features available from initialization through distillation. Moreover, we remark that its performance gain is consistent in all settings, further attesting to the resilience of the proposed approach.
\section{Ablation Studies}
\label{sec:ablas}
\subsubsection{Breakdown of the individual terms of \methnam.}
\begin{table}[t]
\centering
\caption{Impact of each loss term and of using no memory buffer on \methnam. Results given in the Class-IL scenario following the same experimental settings as Tab.\ref{table:c10}-\ref{table:cub200}.}
\begin{tabular}{ccc@{\hskip 0.2cm}ccc@{\hskip 0.2cm}ccc@{\hskip 0.2cm}ccc}
\hline
\noalign{\smallskip}
$~\loss{CL}~$ & $~\loss{FP}~$ & $\loss{AUX}$ & \multicolumn{3}{c@{\hskip 0.2cm}}{\textbf{Split CIFAR-10}} & \multicolumn{3}{c@{\hskip 0.2cm}}{\textbf{\textbf{Split CIFAR-100}}} & \multicolumn{3}{c@{\hskip 0.2cm}}{\textbf{Split CUB-200}}\\
\noalign{\smallskip}
\hline
\noalign{\smallskip}
\multicolumn{3}{c@{\hskip 0.2cm}}{\textbf{Buffer Size}} &
$\nicefrac{\text{w/o}}{\text{buf.}}$ & 500 & 5120 & $\nicefrac{\text{w/o}}{\text{buf.}}$ & 500 & 2000 & $\nicefrac{\text{w/o}}{\text{buf.}}$ & 400 & 1000\\
\noalign{\smallskip}
\hline
\noalign{\smallskip}
\cmark & \cmark & \cmark  & $-$ & \textbf{\rebres{83.65}} & \textbf{\rebres{89.55}} & $-$ & \textbf{\rebres{56.83}} & \textbf{\rebres{64.46}} & $-$ & \textbf{\rebres{59.67}} & \textbf{\rebres{68.32}} \\
\cmark & \xmark & \xmark & $-$ & \rebres{75.79} & \rebres{87.54} & $-$ & \rebres{44.01} & \rebres{57.84} & $-$ & \rebres{56.53} & \rebres{67.29} \\
\cmark & \cmark & \xmark & $-$ & \underline{\rebres{83.29}} & \underline{\rebres{89.53}} & $-$ & \underline{\rebres{55.50}} & \underline{\rebres{63.53}} & $-$ & \underline{\rebres{59.06}}& \underline{\rebres{67.83}}\\
\xmark & \cmark & \xmark &\rebres{60.07} & \rebres{62.63} & \rebres{62.75} & \rebres{49.14} & \rebres{50.20} & \rebres{50.22} & \rebres{37.57} & \rebres{38.43} & \rebres{38.93} \\
\xmark & \cmark & \cmark &\rebres{60.90} & \rebres{63.19} &\rebres{63.79}  & \rebres{49.74} & \rebres{50.88} & \rebres{50.52} & \rebres{37.99} & \rebres{39.20} & \rebres{39.31} \\
\noalign{\smallskip}
\hline
\end{tabular}
\label{tab:loss_ablation}
\end{table}

To better understand the importance of the distinct loss terms in Eq.~\ref{eq:overall} and their connection, we explore their individual contribution to the final accuracy of TwF in Tab.~\ref{tab:loss_ablation}. Based on these results, we make the following observations: \textit{i)} $\loss{CL}$ is the most influential loss term and it is indispensable to achieve results in line with the SOTA; \textit{ii)} $\loss{FP}$ applied on top of $\loss{CL}$ induces better handling of pretraining transfer, as testified by the increased accuracy; \textit{iii)} $\loss{AUX}$ on top of $\loss{FP}$ reduces activation overlapping and brings a small but consistent improvement.

Further, in the columns labeled as $\nicefrac{\text{w/o}}{\text{buf.}}$, we consider what happens if \methnam is allowed \textbf{no replay example at all} and only optimizes $\loss{FP}$ and $\loss{AUX}$ on current task examples. Compared to oEwC in Tab.~\ref{table:c10}-\ref{table:cub200} -- the best non-replay method in our experiments -- we clearly see preserving pretraining features is in itself a much more effective approach, even with rehearsal is out of the picture. 
\subsubsection{Alternatives for the preservation of pretraining knowledge.} \methnam is designed to both preserve pretraining knowledge and facilitate its transfer. However, other approaches could be envisioned for the same purpose. Hence, we compare here \methnam with two alternative baselines for pretraining preservation.
\subsubsection{Pretraining preservation with EwC.}
\begin{figure}[t]
    \centering
    \includegraphics[width=\linewidth]{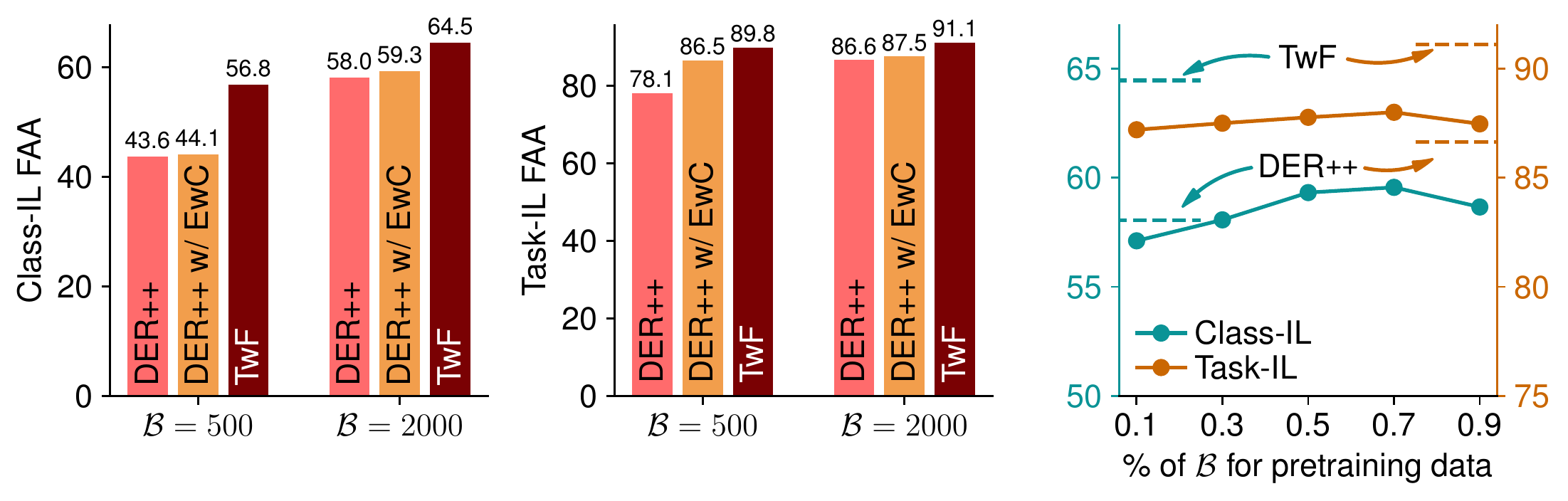}
    \caption{Class-IL (left) and Task-IL (center) FAA performance comparison of our proposal with different possible methods to retain knowledge from pretrain. (Right) Influence of different allocation rates of pretrain examples in $\mathcal{B}$ for \dpp, $|\mathcal{B}|=2000$.}
    \label{fig:ablation_ewc}
\end{figure}
We complement a strong approach such as \dpp with an additional regularization term based on EwC:
\begin{equation}
    \label{eq:objew}
    \mathcal{L}_\text{EwC} = \lambda (\theta - \theta^t)^T \operatorname{diag}(F) (\theta - \theta^t),
\end{equation}
where $\operatorname{diag}(F)$ indicates the diagonal of the empirical Fisher Information Matrix, estimated on the pretraining data at the optimum $\theta^t$. When equipped with this additional loss term, \dpp is anchored to its initialization and prevented from changing its pretraining weights significantly, while its replay-based loss term prevents forgetting of knowledge acquired in previous tasks. As shown by Fig.~\ref{fig:ablation_ewc} (left, center), the EwC loss allows \dpp to improve its accuracy on Split CIFAR-100 with Tiny ImageNet pretraining (especially in the Task-IL setting). However, this improvement is not actively incentivizing feature reuse and thus falls short of \methnam. We finally remark that \methnam and \dpp w/ EwC have a comparable memory footprint (both retain the initialization checkpoint).

\subsubsection{Pretraining preservation through rehearsal.} An alternative for preserving the source knowledge is to assume that pretraining data is available and can be treated as an auxiliary data stream~\cite{bellitto2022effects}. To evaluate this strategy with a bounded memory footprint, we test our baseline method (\dpp) on Split CIFAR-100 with different percentages of the buffer dedicated to pretraining images (from Tiny ImageNet). The results shown in Fig.~\ref{fig:ablation_ewc} (right) confirm our main claim: \dpp coupled with pretraining rehearsal improves over \dpp with only pretraining. This finding proves that, if pretraining is available, it is beneficial to guard it against catastrophic forgetting. 

Furthermore, we highlight that \methnam outperforms the baseline introduced here. When replaying pretraining data, indeed, the model has to maintain its predictive capabilities on the classes of the source task, \textit{i.e.}, we enforce both backward and forward transfer. \methnam, instead, allows the model to disregard the classes of the source dataset, as long as the transfer of its internal representations favors the learning of new tasks ($\Rightarrow$ \textbf{it only enforces forward transfer}). This substantial distinction helps to understand the merits of \methnam: namely, a full but still functional exploitation of the pretraining knowledge.
\subsubsection{Role of pretraining datasets.}
\begin{table}[t]
\begin{center}
\caption{Dissimilar pretrain tasks: accuracy on CIFAR-100 pretrained on SVHN.}
\label{table:ablat_svhn}
\begin{tabular}{l@{\hskip 0.5cm}cc@{\hskip 0.5cm}cc}
\noalign{\smallskip}
\hline
\noalign{\smallskip}
    \textbf{FAA (FF)} & \multicolumn{2}{c@{\hskip 0.5cm}}{\textbf{Class-IL}} & \multicolumn{2}{c}{\textbf{Task-IL}} \\
    \noalign{\smallskip}
\hline
\noalign{\smallskip}
    \textbf{Buffer size} & 500 & 2000 & 500 & 2000 \\
    \noalign{\smallskip}
\hline
\noalign{\smallskip}
    iCaRL~\cite{rebuffi2017icarl} & \normres{39.59}{21.81} & \normres{42.02}{18.78} & \normres{78.89}{4.04} & \normres{80.65}{2.24} \\
    \dpp~\cite{buzzega2020dark}     & \normres{36.46}{53.47} & \normres{52.29}{24.04} & \normres{75.05}{16.22} & \normres{83.36}{8.04} \\
    \textbf{\methnam (ours)}        & \boldres{43.56}{40.02} & \boldres{56.15}{21.51} & \boldres{80.89}{10.12} & \boldres{87.30}{3.12} \\
    \noalign{\smallskip}
\hline
\noalign{\smallskip}
\end{tabular}
\end{center}
\end{table}
Here, we seek to gain further proof of our claim about the ability of \methnam to adapt features from the pretrain. Specifically, we study a scenario where the source data distribution and the target one are highly dissimilar: namely, we first pretrain a ResNet18 backbone on SVHN~\cite{netzer2011reading} and then follow with Split CIFAR-100. We compare our model with the second-best performer from Tab.~\ref{table:c100}, \textit{i.e.},\ iCaRL, and \dpp. The results, reported in Tab.~\ref{table:ablat_svhn}, suggest that our method outranks the competitors not only when pretrained on a similar dataset -- as in Tab.~\ref{table:c100} -- but also when the tasks are very dissimilar. We argue that this result further shows the ability of \methnam to identify which pretraining features are really advantageous to transfer.
\section{Conclusions}
We introduced \methodname, a hybrid method combining Rehearsal and Feature transfer, designed to exploit pretrained weights in an incremental scenario. It encourages feature sharing throughout all tasks, yielding a stable performance gain across multiple settings. We also show that \methnam outperforms other hybrid methods based on rehearsal and regularization and that it is able to profit even from pretraining on a largely dissimilar dataset.

\subsubsection{Acknowledgments.} This paper has been supported from Italian Ministerial grant PRIN 2020 ``LEGO.AI: LEarning the Geometry of knOwledge in AI systems'', n. 2020TA3K9N. Matteo Pennisi is a PhD student enrolled in the National PhD in Artificial Intelligence, XXXVII cycle, course on Health and life sciences, organized by Università Campus Bio-Medico di Roma.

\clearpage
%
%
\bibliographystyle{splncs04}
\bibliography{meta/bib_headers/full,meta/bibliography}

\clearpage
\appendix

\section{Additional Details on the Model}

In this section, we report some additional details on the inner workings of the model which were omitted in the main paper for the sake of brevity.

\subsection{Further details on $\mathbb{M}_\text{Sp}$}
The spatial attention map $\mathbb{M}_\text{Sp}$ is computed on top of the activations of a given layer of the fixed sibling network $\widehat{h} \in \mathbb{R}^{b \times c \times h \times w}$, processed through a ResNet-inspired bottleneck structure~\cite{he2016deep,park2018bam}. In detail, we expand and detail Eq.~\ref{eq:ms} in the main paper:
\begin{equation}
    \label{eq:msp}
    \mathbb{M}_{\operatorname{Sp}} \triangleq \operatorname{C}_{1\times1}^\text{C} \circ \operatorname{ReLU} \circ \operatorname{BN} \circ \operatorname{C}_{3\times3}^\text{B} \circ \operatorname{ReLU} \circ \operatorname{BN} \circ \operatorname{C}_{3\times3}^\text{B} \circ \operatorname{ReLU} \circ \operatorname{BN} \circ \operatorname{C}_{1\times1}^\text{A},
\end{equation}
where $\operatorname{ReLU}$ denotes a ReLU activation, $\operatorname{BN}$ indicates a Batch Normalization layer (conditioned on the task-identifier) and $\operatorname{C}$ indicates a Convolutional layer. More specifically, $\operatorname{C}_{1\times1}^\text{A}$ is a $1 \times 1$ convolution, projecting from $c$ channels to $\nicefrac{c}{4}$; $\operatorname{C}_{3\times3}^\text{B}$ is a $3 \times 3$ dilated convolution with dilation factor $2$ and adequate padding to maintain the same spatial resolution as the input, with $\nicefrac{c}{4}$ channels both as input and output; $\operatorname{C}_{1\times1}^\text{C}$ is a $1 \times 1$ convolution projecting from $\nicefrac{c}{4}$ channels to $1$ channel. This results in $\mathbb{M}_{\operatorname{Sp}}$ having shape $b \times 1 \times h \times w$.

\subsection{Scaling of $\mathbb{M}$}


The second distillation term in Eq.~\ref{eq:obj_2b} requires storing the binary attention maps $\mathbb{M}$ computed for each sample stored in the memory buffer. While this implies a memory overhead, we point out that this is limited by two factors:
\begin{itemize}
    \item The binary nature of $\mathbb{M}$ means its elements can be saved using the smallest supported data-type (usually $1$ byte due to hardware constraints);
    \item As $\mathbb{M}$ usually encodes low level features, it contains several redundancies that can be exploited by (a) using lossless compression algorithms, or (b) down-sampling its spatial dimensions before saving. 
\end{itemize}

In \methnam we save the feature maps $\mathbb{M}$ as bytes and apply down-scaling -- with \textit{nearest neighbor} rule -- with a factor of $2$ if the spatial dimensions are over $16\times 16$. We use the same strategy to up-scale the maps before computing Eq.~\ref{eq:obj_2b}. 

\section{Hyperparameters}
For the experiments of Sec.~\ref{sec:exp}, we employed a choice of hyperparameters validated by grid-search on a random split of $10\%$ of the training set. In the following, we list the values resulting from this process, which can be used to replicate our result. For the sake of fairness, we initialize all models from the same pre-training weights and fix the allowance in terms of iterations and sample efficiency by excluding the number of epochs, lr decay schedule and batch size from the grid-search\footnote{It must be noted that, to allow for its regular operation, CO$^2$L demands a larger batch size. All results for this method are influenced by this advantage.}.

\begin{center}

    \resizebox{.42\paperheight}{!}{%
    \begin{tabular}{ccl}
    \noalign{\smallskip}
    \hline
    \noalign{\smallskip}

\multicolumn{3}{c}{\textbf{Split CIFAR-10 - Class-IL}}\\\noalign{\smallskip}
    \hline
    \noalign{\smallskip}
\textit{shared} & &  $\text{Eps}: 50$ $\text{bs}: 32$ $\text{Eps}_\text{pretr}: 200$ $\text{lr}_\text{decay}: \text{no}$\\
JOINT & - &$\text{lr}:0.1$ \\
SGD & - &$\text{lr}:0.1$ \\
oEwC & - &$\text{lr}:0.1$ $\lambda:10$ $\gamma:1$ \\
LwF & - &$\text{lr}:0.1$ $\alpha:0.3$ $\tau:2$ $\text{wd}:0.0001$ \\
ER & $500$ &$\text{lr}:0.1$ \\
  & $5120$ &$\text{lr}:0.1$ \\
CO$^2$L & $500$ &$\text{lr}:0.5$ $\text{bs}: 256$ $\kappa:0.2$ $\lambda:1$ $\text{lr}_\text{lin}:1$ $\text{lr}^\text{lin}_\text{decay}:0.2$ $\kappa^*:0.01$ $\tau:0.07$ \\
  & $5120$ &$\text{lr}:0.5$ $\text{bs}: 256$ $\kappa:0.2$ $\lambda:1$ $\text{lr}_\text{lin}:1$ $\text{lr}^\text{lin}_\text{decay}:0.2$ $\kappa^*:0.01$ $\tau:0.07$ \\
iCaRL & $500$ &$\text{lr}:0.1$ $\text{wd}:10^{-5}$ \\
  & $5120$ &$\text{lr}:0.03$ $\text{wd}:10^{-5}$ \\
\dpp & $500$ &$\text{lr}:0.03$ $\alpha:0.2$ $\beta:0.5$ \\
  & $5120$ &$\text{lr}:0.03$ $\alpha:0.1$ $\beta:1$ \\
ER-ACE & $500$ &$\text{lr}:0.03$ \\
  & $5120$ &$\text{lr}:0.03$ \\
\methnam & $500$ &$\text{lr}:0.03$ $\alpha:0.3$ $\beta:0.9$ $\lambda:0.1$ $\lambda_\text{FP}:5 \times 10^{-3}$ $\lambda_{\text{FP}}^\text{repl}:0.1$ \\
  & $5120$ &$\text{lr}:0.1$ $\alpha:0.3$ $\beta:0.9$ $\lambda:0.1$ $\lambda_\text{FP}:5 \times 10^{-3}$ $\lambda_{\text{FP}}^\text{repl}:0.3$ \\
\noalign{\smallskip}
    \hline
    \noalign{\smallskip}
    \end{tabular}
    
    }
    
    \resizebox{.42\paperheight}{!}{%
    \begin{tabular}{ccl}
\noalign{\smallskip}
    \hline
    \noalign{\smallskip}
    
\multicolumn{3}{c}{\textbf{Split CIFAR-10 - Task-IL}}\\\noalign{\smallskip}
    \hline
    \noalign{\smallskip}
\textit{shared} &&  $\text{Eps}: 50$ $\text{bs}: 32$ $\text{Eps}_\text{pretr}: 200$ $\text{lr}_\text{decay}: \text{no}$\\
JOINT & - &$\text{lr}:0.1$ \\
SGD & - &$\text{lr}:0.1$ \\
oEwC & - &$\text{lr}:0.03$ $\lambda:0.5$ $\gamma:1$ \\
LwF & - &$\text{lr}:0.01$ $\alpha:0.3$ $\tau:2$ $\text{wd}:0.0001$ \\
ER & $500$ &$\text{lr}:0.1$ \\
  & $5120$ &$\text{lr}:0.1$ \\
CO$^2$L & $500$ &$\text{lr}:0.5$ $\text{bs}: 256$ $\kappa:0.2$ $\lambda:1$ $\text{lr}_\text{lin}:1$ $\text{lr}^\text{lin}_\text{decay}:0.2$ $\kappa^*:0.01$ $\tau:0.07$ \\
  & $5120$ &$\text{lr}:0.5$ $\text{bs}: 256$ $\kappa:0.2$ $\lambda:1$ $\text{lr}_\text{lin}:1$ $\text{lr}^\text{lin}_\text{decay}:0.2$ $\kappa^*:0.01$ $\tau:0.07$ \\
iCaRL & $500$ &$\text{lr}:0.1$ $\text{wd}:10^{-5}$ \\
  & $5120$ &$\text{lr}:0.03$ $\text{wd}:10^{-5}$ \\
\dpp & $500$ &$\text{lr}:0.03$ $\alpha:0.2$ $\beta:0.5$ \\
  & $5120$ &$\text{lr}:0.03$ $\alpha:0.1$ $\beta:1$ \\
ER-ACE & $500$ &$\text{lr}:0.03$ \\
  & $5120$ &$\text{lr}:0.03$ \\
\methnam & $500$ &$\text{lr}:0.03$ $\alpha:0.3$ $\beta:0.9$ $\lambda:0.1$ $\lambda_\text{FP}:5 \times 10^{-3}$ $\lambda_{\text{FP}}^\text{repl}:0.1$ \\
  & $5120$ &$\text{lr}:0.1$ $\alpha:0.3$ $\beta:0.9$ $\lambda:0.1$ $\lambda_\text{FP}:5 \times 10^{-3}$ $\lambda_{\text{FP}}^\text{repl}:0.3$ \\
\noalign{\smallskip}
    \hline
    \noalign{\smallskip}

    \end{tabular}
    
    }

    \resizebox{.45\paperheight}{!}{%
    \begin{tabular}{ccl}
    \noalign{\smallskip}
    \hline
    \noalign{\smallskip}

\multicolumn{3}{c}{\textbf{Split CIFAR-100 - Class-IL}}\\\noalign{\smallskip}
    \hline
    \noalign{\smallskip}
\textit{shared} & &  $\text{Eps}: 50$ $\text{bs}: 64$ $\text{Eps}_\text{pretr}: 200$ $\text{lr}_\text{decay}: 0.1$ $\text{lr}_\text{decay}^\text{steps}: [35, 45]$\\
JOINT & - &$\text{lr}:0.1$ \\
SGD & - &$\text{lr}:0.1$ \\
oEwC & - &$\text{lr}:0.1$ $\lambda:5$ $\gamma:1$ \\
LwF & - &$\text{lr}:0.03$ $\alpha:0.3$ $\tau:2$ $\text{wd}:0.0005$ \\
ER & $500$ &$\text{lr}:0.01$ \\
  & $2000$ &$\text{lr}:0.01$ \\
CO$^2$L & $500$ &$\text{lr}:0.1$ $\text{bs}: 256$ $\kappa:0.2$ $\lambda:1$ $\text{lr}_\text{lin}:1$ $\text{lr}^\text{lin}_\text{decay}:0.2$ $\kappa^*:0.01$ $\tau:0.07$ \\
  & $2000$ &$\text{lr}:0.1$ $\text{bs}: 256$ $\kappa:0.2$ $\lambda:1$ $\text{lr}_\text{lin}:1$ $\text{lr}^\text{lin}_\text{decay}:0.2$ $\kappa^*:0.01$ $\tau:0.07$ \\
iCaRL & $500$ &$\text{lr}:1$ $\text{wd}:10^{-5}$ \\
  & $2000$ &$\text{lr}:1$ $\text{wd}:10^{-5}$ \\
\dpp & $500$ &$\text{lr}:0.1$ $\alpha:0.3$ $\beta:0.3$ \\
  & $2000$ &$\text{lr}:0.1$ $\alpha:0.1$ $\beta:0.5$ \\
ER-ACE & $500$ &$\text{lr}:0.1$ \\
  & $2000$ &$\text{lr}:0.1$ \\
\methnam & $500$ &$\text{lr}:0.03$ $\alpha:0.3$ $\beta:1.2$ $\lambda:0.3$ $\lambda_\text{FP}:0.03$ $\lambda_{\text{FP}}^\text{repl}:1.5$ \\
  & $2000$ &$\text{lr}:0.1$ $\alpha:0.3$ $\beta:1.2$ $\lambda:0.3$ $\lambda_\text{FP}:5 \times 10^{-3}$ $\lambda_{\text{FP}}^\text{repl}:1.2$ \\
\noalign{\smallskip}
    \hline
    \noalign{\smallskip}
    \end{tabular}
    
    }
    
    \resizebox{.45\paperheight}{!}{%
    \begin{tabular}{ccl}
\noalign{\smallskip}
    \hline
    \noalign{\smallskip}
    
\multicolumn{3}{c}{\textbf{Split CIFAR-100 - Task-IL}}\\\noalign{\smallskip}
    \hline
    \noalign{\smallskip}
\textit{shared} & &  $\text{Eps}: 50$ $\text{bs}: 64$ $\text{Eps}_\text{pretr}: 200$ $\text{lr}_\text{decay}: 0.1$ $\text{lr}_\text{decay}^\text{steps}: [35, 45]$\\
JOINT & - &$\text{lr}:0.1$ \\
SGD & - &$\text{lr}:0.01$ \\
oEwC & - &$\text{lr}:0.01$ $\lambda:0.5$ $\gamma:0.7$ \\
LwF & - &$\text{lr}:0.03$ $\alpha:0.3$ $\tau:2$ $\text{wd}:0.0005$ \\
ER & $500$ &$\text{lr}:0.01$ \\
  & $2000$ &$\text{lr}:0.01$ \\
CO$^2$L & $500$ &$\text{lr}:0.1$ $\text{bs}: 256$ $\kappa:0.2$ $\lambda:1$ $\text{lr}_\text{lin}:1$ $\text{lr}^\text{lin}_\text{decay}:0.2$ $\kappa^*:0.01$ $\tau:0.07$ \\
  & $2000$ &$\text{lr}:0.1$ $\text{bs}: 256$ $\kappa:0.2$ $\lambda:1$ $\text{lr}_\text{lin}:1$ $\text{lr}^\text{lin}_\text{decay}:0.2$ $\kappa^*:0.01$ $\tau:0.07$ \\
iCaRL & $500$ &$\text{lr}:1$ $\text{wd}:10^{-5}$ \\
  & $2000$ &$\text{lr}:1$ $\text{wd}:10^{-5}$ \\
\dpp & $500$ &$\text{lr}:0.1$ $\alpha:0.3$ $\beta:1.2$ \\
  & $2000$ &$\text{lr}:0.1$ $\alpha:0.1$ $\beta:0.5$ \\
ER-ACE & $500$ &$\text{lr}:0.1$ \\
  & $2000$ &$\text{lr}:0.1$ \\
\methnam & $500$ &$\text{lr}:0.03$ $\alpha:0.3$ $\beta:1.2$ $\lambda:0.3$ $\lambda_\text{FP}:0.03$ $\lambda_{\text{FP}}^\text{repl}:1.5$ \\
  & $2000$ &$\text{lr}:0.1$ $\alpha:0.3$ $\beta:0.8$ $\lambda:0.3$ $\lambda_\text{FP}:0.03$ $\lambda_{\text{FP}}^\text{repl}:0.3$ \\
\noalign{\smallskip}
    \hline
    \noalign{\smallskip}

    \end{tabular}
    }

    \resizebox{.45\paperheight}{!}{%
    \begin{tabular}{ccl}
    \noalign{\smallskip}
    \hline
    \noalign{\smallskip}

\multicolumn{3}{c}{\textbf{Split CUB-200 - Class-IL}}\\\noalign{\smallskip}
    \hline
    \noalign{\smallskip}
\textit{shared} & &  $\text{Eps}: 50$ $\text{bs}: 64$ $\text{Eps}_\text{pretr}: 50$\\
JOINT & - &$\text{lr}:0.1$ \\
SGD & - &$\text{lr}:0.1$ \\
oEwC & - &$\text{lr}:0.01$ $\lambda:1$ $\gamma:1$ \\
LwF & - &$\text{lr}:0.1$ $\alpha:1$ $\tau:2$ $\text{wd}:0.0005$ \\
ER & $400$ &$\text{lr}:0.03$ \\
  & $1000$ &$\text{lr}:0.1$ \\
CO$^2$L & $400$ &$\text{lr}:0.1$ $\text{bs}: 256$ $\kappa:0.2$ $\lambda:1$ $\text{lr}_\text{lin}:1$ $\text{lr}^\text{lin}_\text{decay}:0.2$ $\kappa^*:0.01$ $\tau:0.07$ \\
  & $1000$ &$\text{lr}:0.1$ $\text{bs}: 256$ $\kappa:0.2$ $\lambda:1$ $\text{lr}_\text{lin}:1$ $\text{lr}^\text{lin}_\text{decay}:0.2$ $\kappa^*:0.01$ $\tau:0.07$ \\
iCaRL & $400$ &$\text{lr}:0.1$ $\text{wd}:10^{-5}$ \\
  & $1000$ &$\text{lr}:0.1$ $\text{wd}:10^{-5}$ \\
\dpp & $400$ &$\text{lr}:0.1$ $\alpha:1$ $\beta:0.5$ \\
  & $1000$ &$\text{lr}:0.1$ $\alpha:0.5$ $\beta:0.5$ \\
  
ER-ACE & $400$ &$\text{lr}:0.1$ \\
  & $1000$ &$\text{lr}:0.1$ \\
\methnam & $400$ &$\text{lr}:0.03$ $\alpha:1$ $\beta:1$ $\lambda:0.3$ $\lambda_\text{FP}:5 \times 10^{-4}$ $\lambda_{\text{FP}}^\text{repl}:0.1$ \\
  & $1000$ &$\text{lr}:0.03$ $\alpha:1$ $\beta:1.2$ $\lambda:0.3$ $\lambda_\text{FP}:5 \times 10^{-4}$ $\lambda_{\text{FP}}^\text{repl}:0.1$ \\
\noalign{\smallskip}
    \hline
    \noalign{\smallskip}
    \end{tabular}
    
    }
    
    \resizebox{.45\paperheight}{!}{%
    \begin{tabular}{ccl}
\noalign{\smallskip}
    \hline
    \noalign{\smallskip}
\multicolumn{3}{c}{\textbf{Split CUB-200 - Task-IL}}\\\noalign{\smallskip}
    \hline
    \noalign{\smallskip}
\textit{shared} & &  $\text{Eps}: 50$ $\text{bs}: 64$ $\text{Eps}_\text{pretr}: 50$\\
JOINT & - &$\text{lr}:0.1$ \\
SGD & - &$\text{lr}:0.1$ \\
oEwC & - &$\text{lr}:0.1$ $\lambda:0.5$ $\gamma:0.9$ \\
LwF & - &$\text{lr}:0.1$ $\alpha:1$ $\tau:2$ $\text{wd}:0.0005$ \\
ER & $400$ &$\text{lr}:0.1$ \\
  & $1000$ &$\text{lr}:0.1$ \\
CO$^2$L & $400$ &$\text{lr}:0.1$ $\text{bs}: 256$ $\kappa:0.2$ $\lambda:1$ $\text{lr}_\text{lin}:1$ $\text{lr}^\text{lin}_\text{decay}:0.2$ $\kappa^*:0.01$ $\tau:0.07$ \\
  & $1000$ &$\text{lr}:0.1$ $\text{bs}: 256$ $\kappa:0.2$ $\lambda:1$ $\text{lr}_\text{lin}:1$ $\text{lr}^\text{lin}_\text{decay}:0.2$ $\kappa^*:0.01$ $\tau:0.07$ \\
iCaRL & $400$ &$\text{lr}:0.1$ $\text{wd}:10^{-5}$ \\
  & $1000$ &$\text{lr}:0.1$ $\text{wd}:10^{-5}$ \\
\dpp & $400$ &$\text{lr}:0.1$ $\alpha:0.5$ $\beta:0.5$ \\
  & $1000$ &$\text{lr}:0.1$ $\alpha:0.5$ $\beta:0.5$ \\
ER-ACE & $400$ &$\text{lr}:0.1$ \\
  & $1000$ &$\text{lr}:0.1$ \\
\methnam & $400$ &$\text{lr}:0.03$ $\alpha:0.3$ $\beta:1$ $\lambda:0.3$ $\lambda_\text{FP}:5 \times 10^{-4}$ $\lambda_{\text{FP}}^\text{repl}:0.1$ \\
  & $1000$ &$\text{lr}:0.03$ $\alpha:1$ $\beta:1$ $\lambda:0.3$ $\lambda_\text{FP}:5 \times 10^{-4}$ $\lambda_{\text{FP}}^\text{repl}:0.1$ \\
\noalign{\smallskip}
    \hline
    \noalign{\smallskip}

    \end{tabular}
    }
\end{center}
\end{document}